\documentclass{article}
\usepackage{spconf,amsmath,graphicx,hyperref}
\usepackage{multirow}    
\usepackage{booktabs}    
\usepackage{hyperref}    
\usepackage{xcolor}      
\usepackage{graphicx}      
\usepackage{tikz}          
\usepackage{pgf-pie}       
\usepackage[T1]{fontenc} 
\usepackage{tabularx}   
\usepackage{ragged2e}   
\usepackage[table]{xcolor}

\usepackage{algorithm}
\usepackage{algorithmic}

\title{GRETEL: A Goal-driven Retrieval and Execution-based Trial Framework for LLM Tool Selection Enhancing}
\name{Zongze Wu, Yani Guo, Churong Liang, Runnan Li$^{\ast}$}
\address{Beijing University of Posts and Telecommunications, Beijing, China}
\begin{document}
%
\maketitle
\begin{abstract}
Despite remarkable advances in Large Language Model capabilities, tool retrieval for agent-based systems remains fundamentally limited by reliance on semantic similarity, which fails to capture functional viability. Current methods often retrieve textually relevant but functionally inoperative tools due to parameter mismatches, authentication failures, and execution constraints—a phenomenon we term the \textit{semantic-functional gap}. We introduce \textbf{GRETEL}, to address this gap through systematic empirical validation. GRETEL implements an agentic workflow that processes semantically retrieved candidates through sandboxed plan-execute-evaluate cycles, generating execution-grounded evidence to distinguish truly functional tools from merely descriptive matches. Our comprehensive evaluation on the ToolBench benchmark demonstrates substantial improvements across all metrics: Pass Rate@10 increases from 0.690 to 0.826, Recall@10 improves from 0.841 to 0.867, and NDCG@10 rises from 0.807 to 0.857. These results establish that execution-based validation provides a more reliable foundation for tool selection than semantic similarity alone, enabling more robust agent performance in real-world applications.
\end{abstract}
\section{Introduction}
\label{sec:intro}
The emergence of Large Language Models (LLMs) such as GPT-4 \cite{openai2023gpt4} and Llama \cite{touvron2023llama} represents a fundamental paradigm shift in artificial intelligence. Despite their capabilities, LLMs remain constrained by their training data distribution, lacking access to real-time information or executable actions.

Addressing this limitation, the research community has pursued augmentation of LLMs with external tools—primarily APIs, to provide these essential capabilities \cite{qin2023toolllm,zhang2024opportunities}. This tool-augmented paradigm transforms LLMs into autonomous agents, yet its efficacy depends critically on a fundamental prerequisite \textbf{tool retrieval}.
Contemporary tool retrieval methodologies predominantly employ semantic similarity measures, wherein user queries are matched against tool descriptions within embedding spaces \cite{qin2023toolllm,liu2025empowering}. While effective for identifying textually similar candidates, this approach exhibits a fundamental \textbf{semantic-functional gap} limitation: tools may demonstrate apparent relevance according to their descriptions while remaining functionally inoperative, resulting in task failure. This gap manifests through multiple mechanisms tools may require parameters absent from the query (e.g., getWeather(zip\_code) when provided only city names); they may fail due to authentication or permission constraints; or retrievers may succumb to semantic ambiguity (e.g., conflating financial Apple APIs with agricultural ones). These phenomena establish that textual relevance constitutes a necessary but insufficient condition for functional utility.

\textbf{GRETEL (A Goal-driven Retrieval and Execution-based Trial Framework for Enhancing LLM Tool Selection)} is thus proposed to address this semantic-functional gap. GRETEL operates on the principle of empirical validation through execution. Rather than accepting semantically ranked lists, GRETEL treats them as testable hypotheses. The framework implements an agentic workflow that systematically evaluates candidate tools through sandboxed plan-execute-evaluate cycles \cite{wang2024chain}. Through analysis of authentic execution feedback, including successful JSON responses and specific error conditions, GRETEL develops empirically grounded assessments of each tool's functional applicability. This process eliminates unsuitable candidates and re-ranks tools based on demonstrated utility rather than textual similarity. The contributions of this work can be summarized:
\begin{enumerate}
\item We formally characterize the semantic-functional gap in tool retrieval, demonstrating its detrimental impact on downstream task performance.
\item We introduce GRETEL, a novel execution-driven agentic framework that validates tool suitability through authentic API feedback mechanisms.
\item We conduct comprehensive experiments on the ToolBench benchmark \cite{qin2023toolllm}, establishing that GRETEL substantially outperforms existing semantic retrievers in identifying functionally viable tools.
\end{enumerate}
This paper positions our approach within the existing research landscape in Section~\ref{sec:related}, formally presents the GRETEL framework architecture and operational mechanisms in Section~\ref{sec:method}, provides comprehensive experimental validation in Section~\ref{sec:experiments}, and concludes key findings and future directions in Section~\ref{sec:conclusion}.

\section{Related Work}
\label{sec:related}

\subsection{Tool-Augmented Large Language Models}
The integration of external tools with LLMs has fundamentally transformed their operational capabilities, enabling autonomous agent behavior \cite{xi2023rise}. Seminal contributions such as Toolformer \cite{schick2023toolformer} demonstrated fine-tuning approaches for API utilization, while prompting frameworks like ReAct \cite{yao2023react} established zero-shot reasoning-action sequences for multi-step tasks. Recent advances include specialized frameworks like API-Bank for financial tool integration, Gorilla \cite{patil2024gorilla} for large-scale API calls, and multi-modal approaches like VisualWebArena \cite{koh2024visualwebarena}. Contemporary surveys \cite{yang2024code} document substantial advances in LLM-based agents across diverse application domains. Our research addresses the fundamental upstream challenge of reliable tool selection, which precedes and determines the success of subsequent reasoning-actions.

\subsection{Semantic-based Tool Retrieval}
Within extensive tool libraries, retrieval constitutes the primary and most critical operation. The prevailing methodology employs information retrieval techniques based on semantic similarity \cite{karpukhin2020dense,reimers2019sentence}. This paradigm encodes tool descriptions and user queries into dense vector representations using models such as Sentence-BERT \cite{reimers2019sentence} or specialized retrieval models like DPR \cite{karpukhin2020dense}, conducting retrieval through vector search mechanisms. ToolLLM introduced ToolBench-IR \cite{qin2023toolllm}, implementing a hybrid retriever integrating sparse (BM25) and dense methodologies.  Recent advances have refined semantic retrieval through improved embedding-based methods \cite{thakur2021beir}, multi-modal retrieval architectures \cite{liu2024multimodal}, and cross-lingual approaches \cite{asai2022matter}. Despite effectiveness in identifying textually relevant candidates, these approaches remain constrained by tool description quality and completeness. We demonstrate that this limitation frequently manifests as a semantic-functional gap, wherein retrieved tools exhibit textual plausibility while remaining practically inoperative.

\subsection{Execution-based Tool Validation and Planning}
Acknowledging limitations of static semantic matching, emerging research has investigated dynamic approaches that leverage execution feedback\cite{tang2023toolalpaca}. Self-correction frameworks such as Reflexion \cite{shinn2023reflexion} enable agents to recover from execution errors through iterative refinement. Feedback-driven methods like CRITIC \cite{gou2024critic} integrate execution results into planning processes. However, these approaches primarily address post-retrieval correction rather than retrieval quality itself. GRETEL distinguishes itself by integrating execution-based validation directly into the retrieval phase, proactively evaluating candidate tool functionality to establish validated, high-precision toolsets prior to primary task execution, thereby preventing downstream failures.

\section{Methodology}
\label{sec:method}

\subsection{Problem Formulation}
Let $\mathcal{T} = \{t_1, t_2, \ldots, t_n\}$ be a collection of available tools, where each tool $t_i$ is characterized by its API specification, parameters, and functionality. Given a user query $q$, a semantic retriever $\mathcal{R}$ produces an initial ranking $R(q) = [t_{r_1}, t_{r_2}, \ldots, t_{r_k}]$ based on textual similarity scores.

The core challenge lies in the semantic-functional gap: tools ranked highly by semantic similarity may be functionally inappropriate due to parameter mismatches, authentication failures, or execution constraints. We formalize this as:
\[
P(\text{functional} \mid \text{semantic}) \ll P(\text{functional})
\]
where $P(\text{functional} \mid \text{semantic})$ represents the probability that a semantically relevant tool is functionally viable.

\begin{figure}[htbp]
\centering
\includegraphics[width=0.8\linewidth]{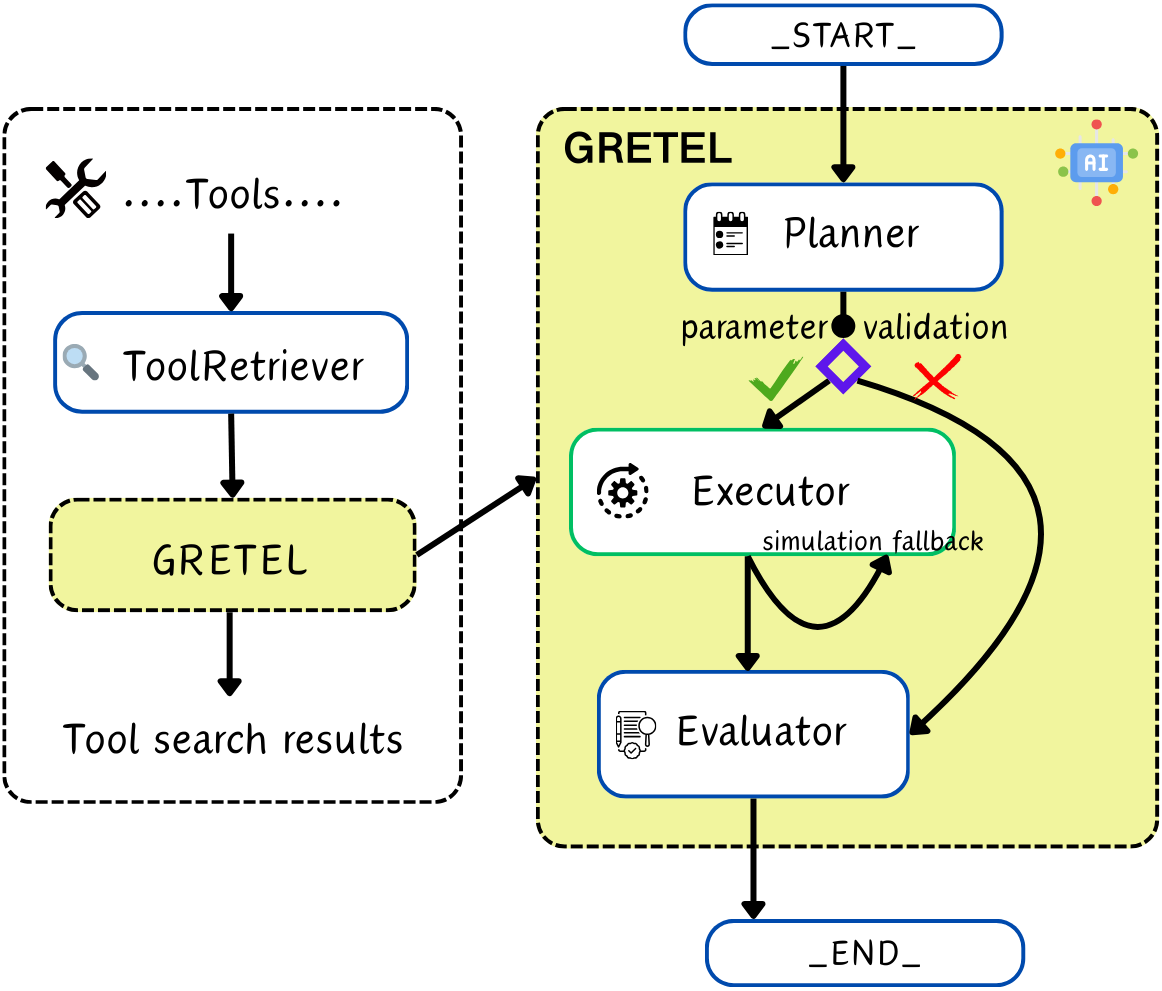}
\caption{GRETEL ingests a semantically retrieved tool list and employs a Plan-Execute-Evaluate loop, yielding a final re-ranking grounded in functional validation.}
\label{fig:framework}
\end{figure}

GRETEL addresses this gap by computing a functionally-grounded re-ranking $R'(q)$ through empirical validation of each candidate tool $t_i \in R(q)$. Rather than passively accepting initial tool rankings, the framework actively evaluates functional viability of candidates through sandboxed, iterative validation processes as shown in Figure~\ref{fig:framework}.

\subsection{The GRETEL Agentic Workflow}
GRETEL implements an agentic workflow through the LangGraph library \cite{langgraph2024}, modeling the process as a stateful graph structure. LangGraph provides orchestration capabilities for multi-agent workflows with enhanced control and reliability \cite{chase2024langgraph}. The workflow systematically processes each candidate tool through a three-stage validation protocol. LLM-powered nodes within this workflow operate under precisely engineered prompts, as detailed in Table~\ref{tab:prompts}.

\begin{table}[t!]
\centering
\caption{Core Prompt Templates for GRETEL}
\label{tab:prompts}
\definecolor{myblue}{rgb}{0.2, 0.4, 0.7}
\definecolor{lightgray}{gray}{0.95} 
\renewcommand{\arraystretch}{1.5}
\resizebox{\columnwidth}{!}{%
\begin{tabular}{
  l 
  >{\RaggedRight}p{7cm} 
  >{\RaggedRight\arraybackslash}p{5.4cm} 
}
\toprule
\textbf{Component} & 
\textbf{Prompt Template (Abbreviated)} & 
\textbf{Purpose} \\
\midrule
\textcolor{myblue}{\textbf{Planner}} & 
\texttt{Given query "\{query\}" and API spec "\{api\_spec\}". Extract parameter values ...} & 
Query-to-API parameter extraction \\
\midrule 
\textcolor{myblue}{\textbf{Simulator}} & 
\texttt{Generate realistic JSON response for failed API call "\{api\_call\}" based on query context...} & 
Plausible response generation for failed executions \\
\midrule
\textcolor{myblue}{\textbf{Evaluator}} & 
\texttt{Rank candidate tools using execution evidence. Output JSON list of [Tool, API] pairs...} & 
Holistic ranking with semantic and execution evidence \\
\bottomrule
\end{tabular}%
}
\end{table}

\subsubsection{Trial-based Evidence Generation}
For candidate tool $t_i$, GRETEL conducts a validation trial to generate functional evidence in two stages with Algorithm~\ref{algo:trial_generation}.

\begin{algorithm}[t!]
\caption{Trial-based Evidence Generation}
\label{algo:trial_generation}
\begin{algorithmic}[1]
\REQUIRE Query $q$, Tool $t_i$ with API specification $\mathcal{A}_i$
\ENSURE Evidence tuple $(status, result, metadata)$
\STATE $params \leftarrow \text{PLAN}(q, \mathcal{A}_i)$ \hfill \textcolor{blue}{\% Planning stage}
\IF{$params = \text{ERROR}$ \OR $params = \text{INVALID\_JSON}$}
    \RETURN $(\textcolor{red}{\text{PLANNING\_FAILED}}, error\_msg, \emptyset)$
\ENDIF
\STATE $api\_call \leftarrow \text{format\_call}(t_i.name, params)$ 
\STATE $result \leftarrow \text{REAL\_EXECUTE}(t_i, params)$ \hfill \textcolor{blue}{\% Real execution}
\IF{$result.status = \text{success}$}
    \STATE $metadata \leftarrow \{simulation\_used: false\}$
\ELSIF{$result.status = \text{error}$}
    \STATE $result \leftarrow \text{SIMULATE}(q, t_i, api\_call, result)$ \hfill \textcolor{teal}{\% LLM simulation}
    \STATE $metadata \leftarrow \{simulation\_used: true\}$ 
    \hfill \textcolor{purple}{\% Track simulation usage}
\ELSE
    \STATE $metadata \leftarrow \{simulation\_used: false\}$
\ENDIF
\RETURN $(result.status, result.data, metadata)$
\end{algorithmic}
\end{algorithm}

\begin{enumerate}
\item \textbf{Planning}. A \textbf{Planner} module uses a LLM, guided by the user query $Q$ and the tool's OpenAPI specification $\mathcal{A}_i$ \cite{openapi2024}, to construct a syntactically valid and semantically plausible API call. A failure at this stage is itself a strong negative signal, indicating the tool's parameters cannot be satisfied by the query.

\item \textbf{Execution}. Successfully generated API calls are dispatched by a sandboxed \textbf{Executor} module. This node captures the real-world outcome, either a successful JSON response or a structured error message like authentication failure. This execution-driven verification provides direct evidence of a tool's practical utility \cite{wang2024execution}. To handle non-critical failures, the Executor can leverage an LLM-based simulation fallback to produce a plausible success response for an otherwise valid tool.
\end{enumerate}

\subsubsection{Holistic Re-ranking with Accumulated Evidence}
The trial process is managed by LangGraph's state machine \cite{langgraph2024}, which iteratively processes each candidate tool and aggregates the resulting evidence. Upon completion of all trials, a final \textbf{Holistic Re-ranker} node receives the original query $Q$ along with the complete set of accumulated evidence, containing the planning and execution outcomes for every candidate. This node prompts an LLM to perform a final, comparative re-ranking of all tools\cite{reimers2019sentence}. The model is instructed to prioritize tools with successful execution or simulation evidence while penalizing those that failed during planning or execution. This holistic, evidence-driven ranking\cite{reimers2019sentence, xu2024enhancing} by a capable LLM replaces rigid scoring, enabling nuanced judgments; formal consistency analysis is left to future work. The final output is the list of tools re-ranked according to this functionally-grounded analysis.

\section{Experiments}
\label{sec:experiments}

Comprehensive experiments are processed to validate GRETEL. The evaluation is designed to validate the two primary research targets:
\textbf{(1)} Does our execution-based re-ranking framework significantly improve upon established semantic tool retrieval baselines on the full benchmark?
\textbf{(2)} What is the contribution of each component within \textsc{GRETEL}, and how does it compare against a strong, LLM-based SOTA method?

\subsection{Experimental Setup}

\noindent \textbf{Dataset}. Our experiments leverage the \textbf{ToolBench} benchmark \cite{qin2023toolllm}, using the full G1 set (~80k queries) for main results and a 10k subset for ablations. While its API categories are diverse, our analysis evaluates overall performance without a breakdown by tool type, a direction left for investigation.

\noindent\textbf{Baselines and Methods}. We compare against PMLM-L3-v2 \cite{PMLM-L3-v2} and ToolBench-IR \cite{qin2023toolllm}. The proposed methods, including GERTEL and its ablated versions,  all build upon ToolBench-IR as a post-processing re-ranking stage.

\noindent\textbf{Evaluation Metrics}. Performance is evaluated using three standard metrics at cutoff points \textbf{K=5,10} for main results and \textbf{K=3,5} for the ablation analysis. The metrics are: \textbf{Recall@K}, \textbf{NDCG@K} (a rank-aware quality measure), and \textbf{Pass Rate@K} (a stringent functional correctness measure).

\subsection{Main Quantitative Results}
Table~\ref{tab:main_results} compares \textsc{GRETEL} against multiple state-of-the-art baselines across four evaluation metrics. Our method demonstrates significant improvements over both classical retrieval approaches and recent execution-aware methods. Compared to the strongest baseline (ToolBench-IR), \textsc{GRETEL} achieves substantial gains in functional correctness: Pass Rate@10 improves from 0.807 to 0.857, while Recall@10 increases from 0.841 to 0.867. Against specialized tool-calling methods like Gorilla, \textsc{GRETEL} showsmore pronounced advantages, with Pass Rate@10 improving from 0.807 to 0.857. The consistent improvements across all metrics validate the proposed approach provides more reliable tool selection than semantic similarity alone.

\begin{table}[t!]
\centering
\caption{Main results on the full ToolBench-I1 dataset.}
\label{tab:main_results}
\resizebox{\columnwidth}{!}{%
\begin{tabular}{lcccccccc}
\toprule
\textbf{Method} & \textbf{Recall@5} & \textbf{Recall@10} & \textbf{NDCG@5} & \textbf{NDCG@10} & \textbf{Pass@5} & \textbf{Pass@10}\\
\midrule
PMLM-L3-v2 & 0.365 & 0.468 & 0.399 & 0.421 & 0.140 & 0.250 \\
ToolkenGPT   & 0.421 & 0.531 & 0.445 & 0.472 & 0.203 & 0.315 \\
ReAct     & 0.389 & 0.492 & 0.412 & 0.441 & 0.165 & 0.278 \\
ToolLLM   & 0.456 & 0.573 & 0.491 & 0.518 & 0.267 & 0.389 \\
ToolBench-IR (Base) & 0.709 & 0.841 & 0.791 & 0.807 & 0.460 & 0.690 \\
\textbf{+ GRETEL (Ours)} & \textbf{0.883} & \textbf{0.867} & \textbf{0.848} & \textbf{0.857} & \textbf{0.658} & \textbf{0.826} \\
\bottomrule
\end{tabular}%
}
\end{table}

\subsection{Ablation and SOTA Analysis}
To dissect the components of GRETEL and situate its performance against a strong contemporary re-ranker, we conducted an analysis on a 10,000-query subset (Table~\ref{tab:ablation_sota}). All methods in test use ToolBench-IR as the initial retriever. We first observe that a standard ``LLM Re-ranker'' provides a formidable SOTA baseline, significantly outperforming the base retriever.

Our ablation study~\cite{chen2025improving}, however, reveals the source of \textsc{GRETEL}'s advantage. Adding direct execution trials (\texttt{+ GRETEL w/o Simulation}) improves the Pass Rate@5 over the SOTA from 0.581 to 0.585, confirming that real-world execution is a more reliable signal than semantic relevance alone. The final addition of our simulation fallback mechanism in the full \textsc{GRETEL} model provides a further boost across all metrics. The result states each component of \textsc{GRETEL} contributes meaningfully for tools retrieval.

\begin{table}[t]
\centering
\caption{Ablation result for components with ToolBench-IR.}
\label{tab:ablation_sota}
\resizebox{\columnwidth}{!}{%
\begin{tabular}{lcccccc}
\toprule
\textbf{Method (applied on ToolBench-IR)} & \textbf{Recall@3} & \textbf{Recall@5} & \textbf{NDCG@3} & \textbf{NDCG@5} & \textbf{Pass@3} & \textbf{Pass@5} \\
\midrule
(No Re-ranker, Base)        & 0.456 & 0.564 & 0.425 & 0.484 & 0.342 & 0.525 \\
+ LLM Re-ranker (SOTA)      & 0.573 & 0.650 & 0.567 & 0.609 & 0.466 & 0.581 \\
+ \textsc{GRETEL} w/o Sim   & 0.572 & 0.648 & 0.564 & 0.606 & 0.452 & 0.585 \\
\textbf{+ \textsc{GRETEL} (Full, Ours)} & \textbf{0.612} & \textbf{0.682} & \textbf{0.603} & \textbf{0.643} & \textbf{0.500} & \textbf{0.629} \\
\bottomrule
\end{tabular}%
}
\end{table}

\subsection{Error Analysis: The Semantic-Functional Gap}

To elucidate \textsc{GRETEL}'s effectiveness mechanisms, we analyzed failure modes of filtered tools, as presented in Table~\ref{tab:error_analysis}. This analysis empirically validates our core semantic-functional gap hypothesis, revealing substantial failure rates among semantically plausible tools during execution. The predominant failure mode constitutes \textbf{Parameter Mismatch}, wherein agents cannot construct valid API calls. And \textbf{Semantic Mismatch}, where tools execute successfully yet return empty or irrelevant responses, and \textbf{Execution Failure} attributable to server-side errors \cite{schick2023toolformer,li2025feedback}.

\begin{table}[htbp]
\centering
\caption{Analysis of trial failures for top-5 candidates from ToolBench-IR: 85\% candidates are functionally flawed.}
\label{tab:error_analysis}
\resizebox{\columnwidth}{!}{%
\begin{tabular}{@{}lcl@{}}
\toprule
\textbf{Failure Type} & \textbf{\%} & \textbf{Description} \\
\midrule
Parameter Mismatch & 42\% & LLM fails to construct a valid API call from the query and documentation. \\
Semantic Mismatch  & 25\% & Tool executes but returns an empty or irrelevant response. \\
Execution Failure  & 18\% & The API call itself fails due to server-side or authentication errors. \\
\textit{Functional Success} & \textit{15\%} & \textit{The tool was successfully executed by the trial agent.} \\
\bottomrule
\end{tabular}%
}
\end{table}

\begin{itemize}
\item \textbf{Parameter Mismatch (42\%)}: The predominant failure mode wherein agents, despite tool documentation guidance, cannot construct valid API calls due to hallucinated parameters, incorrect data types, or missing mandatory fields uninferable from queries.
\item \textbf{Semantic Mismatch (25\%)}: Tools execute successfully yet return empty or irrelevant responses. For instance, flight search APIs return no results for valid routes, indicating coverage limitations for airlines or regions nuances overlooked by semantic retrieval.
\item \textbf{Execution Failure (18\%)}: Failures arising from server-side errors, authentication issues, or endpoints that deviate from their documentation.
\end{itemize}
This breakdown empirically validates our central hypothesis regarding the semantic-functional gap in tool retrieval. \textsc{GRETEL} addresses this gap through identification and penalization of these failure modes, ensuring final rankings reflect both semantic relevance and functional robustness.

\subsection{Case Study: Flight Booking Scenario}

To demonstrate \textsc{GRETEL}'s operational mechanisms concretely, a specific query \textit{``Find me a one-way flight from San Francisco to New York for next Tuesday.''} is given in Table~\ref{tab:case_study}.

\begin{table}[thbp]
\centering
\caption{Case study: \textsc{GRETEL} promotes a valid tool over a semantically similar but non-functional alternative.}
\label{tab:case_study}
\resizebox{\columnwidth}{!}{%
\begin{tabular}{@{}cll@{}}
\toprule
\textbf{Rank} & \textbf{Tool \& API} & \textbf{Justification / \textsc{GRETEL} Trial Result} \\
\midrule
\multicolumn{3}{@{}l}{\textit{\textbf{Initial Ranking from ToolBench-IR}}} \\
1 & FlightsPro.search        & High semantic similarity with \emph{flight}, \emph{search}. \\
2 & Kayak.search\_flights    & Good semantic match. \\
3 & Skyscanner.get\_flights  & Relevant keywords. \\
\midrule
\multicolumn{3}{@{}l}{\textit{\textbf{Final Ranking from \textsc{GRETEL}}}} \\
\textbf{1} & \textbf{Kayak.search\_flights}   & \textbf{\textcolor{green!50!black}{SUCCESS}}: Agent built \texttt{from=SFO, to=JFK, date=\dots} and received valid results. \\
2          & \textbf{Skyscanner.get\_flights} & \textbf{\textcolor{green!50!black}{SUCCESS}}: Trial succeeded, ranked lower due to higher latency. \\
\textellipsis & \textellipsis & \textellipsis \\
\textbf{NA} & \textbf{FlightsPro.search}      & \textbf{\textcolor{red!80!black}{FAILURE}}: Missing mandatory \texttt{carrier\_code} parameter in query; demoted by trial evidence. \\
\bottomrule
\end{tabular}%
}
\end{table}

\section{Conclusion}
\label{sec:conclusion}

This work tackles the prevalent semantic-functional gap in tool retrieval with GRETEL, a framework that re-ranks candidates based on direct evidence from trial-based execution. Our empirical results confirm this approach significantly and consistently improves functional correctness (Pass Rate) and rank-aware quality (NDCG). Future work must address its current scope on stateless APIs and its scalability, mitigating the significant computational overhead via optimizations like parallelization and caching. We contend that this crucial insight—prioritizing functional viability over semantic recall—means dynamic validation is not merely an enhancement but a fundamental requirement for building robust and autonomous AI agents that can operate reliably and predictably in complex, real-world environments.

\vfill\pagebreak
\bibliographystyle{IEEEbib}
\bibliography{strings,refs.bib}

\begin{thebibliography}{10}

\bibitem{openai2023gpt4}
OpenAI, Josh Achiam, et~al.,
\newblock ``Gpt-4 technical report,''
\newblock 2023.

\bibitem{touvron2023llama}
Hugo Touvron, Thibaut Lavril, et~al.,
\newblock ``Llama: Open and efficient foundation language models,''
\newblock 2023.

\bibitem{qin2023toolllm}
Yujia Qin, Shihao Liang, et~al.,
\newblock ``Toolllm: Facilitating large language models to master 16000+
  real-world apis,''
\newblock {\em ICLR}, 2024.

\bibitem{zhang2024opportunities}
Marcus~M. Noack, Harinarayan Krishnan, Stephanie Brewer, et~al.,
\newblock ``Opportunities for retrieval and tool augmented large language
  models in scientific facilities,''
\newblock {\em npj Computational Materials}, vol. 10, no. 1, pp. 262, 2024.

\bibitem{liu2025empowering}
Wei Liu, Yuxiang Zhang, et~al.,
\newblock ``Empowering llms by hybrid retrieval-augmented generation for
  domain-centric q\&a in smart manufacturing,''
\newblock {\em Journal of Manufacturing Systems}, vol. 78, pp. 268--281, 2025.

\bibitem{wang2024chain}
Yusen Zhang, Ruoxi Sun, Yanfei Chen, Tomas Pfister, Rui Zhang, and Sercan~Ö.
  Arık,
\newblock ``Chain of agents: Large language models collaborating on
  long-context tasks,''
\newblock 2024, vol.~37, NeurIPS.

\bibitem{xi2023rise}
Zhiheng Xi, Wenxiang Chen, et~al.,
\newblock ``The rise and potential of large language model based agents: A
  survey,''
\newblock 2023.

\bibitem{schick2023toolformer}
Timo Schick, Jane Dwivedi-Yu, et~al.,
\newblock ``Toolformer: Language models can teach themselves to use tools,''
\newblock in {\em Advances in Neural Information Processing Systems}. 2023,
  vol.~36, pp. 68539--68551, NeurIPS.

\bibitem{yao2023react}
Shunyu Yao, , et~al.,
\newblock ``{ReAct}: Synergizing reasoning and acting in language models,''
\newblock in {\em International Conference on Learning Representations (ICLR)},
  2023.

\bibitem{patil2024gorilla}
Shishir~G. Patil et~al.,
\newblock ``Gorilla: Large language model connected with massive {APIs},''
\newblock in {\em Advances in Neural Information Processing Systems (NeurIPS)},
  2024.

\bibitem{koh2024visualwebarena}
Jing~Yu Koh et~al.,
\newblock ``Visualwebarena: Evaluating multimodal agents on realistic visual
  web tasks,''
\newblock Bangkok, Thailand, 2024, pp. 881--905, Association for Computational
  Linguistics.

\bibitem{yang2024code}
Ke~Yang, Jiateng Liu, et~al.,
\newblock ``If llm is the wizard, then code is the wand: A survey on how code
  empowers large language models to serve as intelligent agents,''
\newblock 2024.

\bibitem{karpukhin2020dense}
Vladimir Karpukhin et~al.,
\newblock ``Dense passage retrieval for open-domain question answering,''
\newblock 2020, pp. 6769--6781, Association for Computational Linguistics.

\bibitem{reimers2019sentence}
Nils Reimers and Iryna Gurevych,
\newblock ``Sentence-bert: Sentence embeddings using siamese bert-networks,''
\newblock Hong Kong, China, 2019, Association for Computational Linguistics.

\bibitem{thakur2021beir}
Nandan Thakur et~al.,
\newblock ``Beir: A heterogeneous benchmark for zero-shot evaluation of
  information retrieval models,''
\newblock 2021, vol.~1, Curran Associates, Inc.

\bibitem{liu2024multimodal}
Yiming Liu et~al.,
\newblock ``Multi-modal tool retrieval for large language models,''
\newblock {\em EMNLP}, 2024.

\bibitem{asai2022matter}
Masanari Ishiyama et~al.,
\newblock ``Impact of free energy of polymers on polymorphism of
  polymer-grafted nanoparticles,''
\newblock {\em Soft Matter}, vol. 18, pp. 6318--6325, 2022.

\bibitem{tang2023toolalpaca}
Qiaoyu Tang et~al.,
\newblock ``Toolalpaca: Generalized tool learning for language models with 3000
  examples,''
\newblock in {\em Thirty-seventh Conference on Neural Information Processing
  Systems}, 2023.

\bibitem{shinn2023reflexion}
Noah Shinn, Federico Cassano, et~al.,
\newblock ``Reflexion: Language agents with verbal reinforcement learning,''
\newblock in {\em Advances in Neural Information Processing Systems}. 2023,
  vol.~36, NeurIPS.

\bibitem{gou2024critic}
Zhibin Gou, Shao, et~al.,
\newblock ``Critic: Large language models can self-correct with
  tool-interactive critiquing,''
\newblock in {\em International Conference on Learning Representations (ICLR)
  (Poster)}, 2024.

\bibitem{langgraph2024}
LangChain,
\newblock ``Langgraph: Multi-agent workflows,''
\newblock 2024.

\bibitem{chase2024langgraph}
Harrison Chase et~al.,
\newblock ``Langgraph platform,''
\newblock 2024.

\bibitem{openapi2024}
OpenAPI Initiative,
\newblock ``Openapi specification v3.1.1,''
\newblock 2024.

\bibitem{wang2024execution}
Yiming Wang et~al.,
\newblock ``Execution-based evaluation for tool-augmented language models,''
\newblock {\em ICLR}, 2024.

\bibitem{xu2024enhancing}
Qiancheng Xu, Yongqi Li, Heming Xia, and Wenjie Li,
\newblock ``Enhancing tool retrieval with iterative feedback from large
  language models,''
\newblock 2024, pp. 9609--9619, Association for Computational Linguistics.

\bibitem{PMLM-L3-v2}
Nils Reimers and Iryna Gurevych,
\newblock ``Sentence-bert: Sentence embeddings using siamese bert-networks,''
\newblock Hong Kong, China, Nov. 2019, pp. 3982--3992, Association for
  Computational Linguistics.

\bibitem{chen2025improving}
Qiao Chen et~al.,
\newblock ``Improving large language model applications in biomedicine with
  retrieval-augmented generation: a systematic review, meta-analysis, and
  clinical development guidelines,''
\newblock {\em Journal of the American Medical Informatics Association}, pp.
  891--904, 2025.

\bibitem{li2025feedback}
Xiaoming Li et~al.,
\newblock ``Feedback-driven tool selection in multi-agent systems,''
\newblock {\em AAAI}, 2025.

\end{thebibliography}

\end{document}